\documentclass{article}
\usepackage{arxiv}
\usepackage{makecell}
\usepackage[utf8]{inputenc} % allow utf-8 input
\usepackage[T1]{fontenc}    % use 8-bit T1 fonts
\usepackage{hyperref}       % hyperlinks
\usepackage{url}            % simple URL typesetting
\usepackage{booktabs}       % professional-quality tables
\usepackage{amsfonts}       % blackboard math symbols
\usepackage{nicefrac}       % compact symbols for 1/2, etc.
\usepackage{microtype}      % microtypography
\usepackage{lipsum}
\usepackage{multirow}
\usepackage{graphicx}
\usepackage{caption}
\usepackage{subcaption}
\usepackage{xspace}
\usepackage{geometry}
\usepackage{amsmath,amssymb,amsfonts,dsfont}
\usepackage[lined,  ruled, linesnumbered]{algorithm2e}
\usepackage{xcolor}
\title{Sample-Efficient, Exploration-Based Policy Optimisation for Routing Problems}
\author{Nasrin Sultana, Jeffrey Chan, Tabinda Sarwar, A. K. Qin}
\begin{document}
\maketitle

\begin{abstract}
%Sample efficiency denotes the amount of experience algorithm needs to generate in an environment (e.g. the number of actions it takes and number of resulting states + rewards it observes) during training in order to reach a certain level of performance.
Model-free deep-reinforcement-based learning algorithms have been applied to a range of COPs~\cite{bello2016neural}~\cite{kool2018attention}~\cite{nazari2018reinforcement}. However, these approaches suffer from two key challenges when applied to combinatorial problems: insufficient exploration and the requirement of many training examples of the search space to achieve reasonable performance. Combinatorial optimisation can be complex, characterised by search spaces with many optimas and large spaces to search and learn. Therefore, a new method is needed to find good solutions that are more efficient by being more sample efficient. This paper presents a new reinforcement learning approach that is based on entropy. In addition, we design an off-policy-based reinforcement learning technique that maximises the expected return and improves the sample efficiency to achieve faster learning during training time. We systematically evaluate our approach on a range of route optimisation tasks typically used to evaluate learning-based optimisation, such as the such as the Travelling Salesman problems (TSP), Capacitated Vehicle Routing Problem (CVRP). In this paper, we show that our model can generalise to various route problems, such as the split-delivery VRP (SDVRP), and compare the performance of our method with that of current state-of-the-art approaches. The Empirical results show that the proposed method can improve on state-of-the-art methods in terms of solution quality and computation time and generalise to problems of different sizes.

%, Multiple Routing with Fixed Fleet Problems (MRPFF) 
%(amount of experience the algorithm generates in a search space to achieve a certain level of performance)
\end{abstract}
% keywords can be removed
\section{Introduction}

Route Optimisation problems are an important class of combinatorial optimisation problems~\cite{van1991handbook}. Route Optimisation problems have many real-life applications, e.g., supply chain management, warehouse layout and logistics, aviation planning, healthcare scheduling and hardware design~\cite{baldacci2010some}. A typical approach to solve these problems involves modelling the problem into a mathematical objectives and then selecting an appropriate solver to optimise the problem at hand. These approaches have been successful; however, it requires expert and domain knowledge, limiting their widespread usage. Also, when the problem instance changes, the searching process often needs to be restarted, and any knowledge gained from solving previous instances is not considered or utilised. Hence, these challenges have raised interest in the deep learning methods to learn black-box solvers for such problems and benefit from previous solving efforts.

Recent state-of-the-art approaches in deep learning have shown success in route optimisation problems~\cite{vinyals2015pointer}~\cite{joshi2019efficient}~\cite{sultana2022learning}, particularly deep-reinforcement-learning-based approaches~\cite{bello2016neural}~\cite{nazari2018reinforcement}~\cite{kool2018attention}~\cite{kwon2020pomo}~\cite{sultana2021learning} that use standard on-policy gradient-based methods (PGM)~\cite{williams1992simple}. However, many issues are encountered when using current on-policy methods for optimising route problems, as the search space for route optimisation problems is large. 

One issue with on-policy deep reinforcement learning methods is that they are notoriously expensive due to their lack of sample efficiency.\footnote{Sample efficiency means the amount of experience that an algorithm needs to generate, e.g. the number of actions it takes and the number of resulting states + rewards it observes during training to reach a certain level of performance.} When we formulate route optimisation problems as reinforcement learning problems, we relate the search space with the environment, where we collect samples. These methods rely on sampling to find the (stochastic) gradient of the optimisation objective for route optimisation problems. Due to the typical difficulty and undulating landscape of combinatorial optimisation problems, gradient estimates can suffer from high variance. To collect a better sample (sample efficiency) in an unknown environment, the model needs to visit states that have not been seen before (exploration). Thus, the methods can be trapped in local minima if an effective strategy does not collect enough samples because of a lack of exploration. One of the crucial factors for the poor sample efficiency of deep reinforcement learning methods~\cite{konda2000actor}~\cite{schulman2017proximal}~\cite{mnih2016asynchronous} is that they use a considerable number of samples per step to estimate gradients, which increases the sample inefficiency~\cite{haarnoja2018soft}. Therefore, a compelling exploration strategy requires finding better samples to optimise the policy to achieve the best objective value (i.e. an exploration strategy needs to be employed to encourage veering off of previous paths). Due to the lack of sample efficiency, extensive adoption of reinforcement learning in real-world domains has remained limited~\cite{wu2017scalable}. Even relatively simple tasks require many data collection steps, and complex tasks, such as route (combinatorial) problems, might need substantially more. Therefore, using on-policy deep reinforcement learning methods optimising route problems is difficult~\cite{van1991handbook}.

Another issue with the current reinforcement-learning-based methods is insufficient exploration to solve route optimisation problems. The lack of exploration often results in early convergence to poor policies because of the lack of experience gathered from the environment/observation space, which in turn challenges the methods performance on route optimisation problems. The premature convergence leads to policies becoming deterministic too quickly. Policy optimisation methods rely on better sample collection.\footnote{Suppose two nearby samples can have very different gradients, making it challenging to optimise the objective of route optimisation. Without effectively exploring the environment, the value of the current samples may not produce the best value to find the best objective.} Therefore, effective exploration requires a better estimation of the gradients that can result from better sampling.

In recent years, some of the most successful algorithms, such as Bello et al.~\cite{bello2016neural}, Nazari et al.~\cite{nazari2018reinforcement} and Kool et al.~\cite{kool2018attention} have been trained using policy-based methods that suffer from a lack of exploration and sample efficiency. To encourage exploration, two new methods have been introduced, namely ERRL~\cite{sultana2021learning} and POMO~\cite{kwon2020pomo}. However, their model suffers sample inefficiency because it involves an on-policy learning mechanism. To the best of our
knowledge, existing learning to optimises literature for combinatorial problems have not studied sample efficiency issues to the best of our knowledge.

Our work is an extension of the ERRL~\cite{sultana2021learning} method, instead of an on-policy method, here, we propose an off-policy method that tends to learn from past samples using experience replay buffers that can provide better sample efficiency.~\cite{haarnoja2017reinforcement} Encourage exploration using manual hyper-parameter tuning, which is presumably handcrafted. Alternatively, we propose an automatic selection of the parameters. 

The main contributions are as follows:

\begin{itemize}
\item We propose a new architecture that we call EPOSE. It offers a maximum entropy model that encourages exploration with sample efficiency using off-policy-based policy gradient methods~\cite{haarnoja2018soft}.

 \item The EPOSE algorithm also offers an automatic gradient-based temperature tuning method, which adjusts the amount of exploration over the visited states to match a target value to encourage exploration, limiting the sample complexity~\cite{christodoulou2019soft}.
 
 \item Extensive experimental results of three routing problems show the effectiveness of using the off-policy gradient method and generalise a wide range of routing problems with different constraints. 

\item More importantly, our model demonstrates comparable optimality results: the traditional state-of-the-art non-learning-based heuristics and outperforms previous reinforcement learning methods.  

\end{itemize}

%We contribute to this line of research for routing problems by introducing \emph{Exploration Based Policy Optimisation with Sample Efficiency (EPOSE)}. EPOSE offers a maximum entropy model that encourages exploration with sample efficiency using off-policy based policy gradient methods~\cite{haarnoja2018soft}. EPOSE also offers an automatic gradient-based temperature tuning method, which adjusts the amount of exploration over the visited states to match a target value to encourage exploration, limiting the sample complexity~\cite{christodoulou2019soft}. 

%Extensive experiment results on three routing problems show the effectiveness of using the off-policy gradient method and generalise a wide range of routing problems with different constraints. More importantly, our model demonstrates comparable optimality results to the traditional state-of-the-art non-learning based heuristics and outperforms previous reinforcement learning methods. 

The rest of the paper is organised as follows: Section~\ref{sec:rw} summarises the related literature. Section~\ref{sec:BG} describes our background. Section~\ref{sec:epose} introduces our approach. Section~\ref{sec:exp} gives the computational results. Finally, conclusions and future works are listed in Section~\ref{sec:conclusion}.

%https://www.quora.com/What-are-the-benefits-of-using-off-policy-learning-over-on-policy-learning-in-reinforcement-learning
\section{Related Works}\label{sec:rw}
%\lipsum[4] See Section \ref{sec:headings}.

In this section, we review some literature that is closely related to our work. As previously discussed, many methods have been developed over the last few years to tackle route optimisation problems using recent advances in deep learning methods. Supervised learning approaches are trained by pairs of problem instances and optimal solutions. Deep reinforcement learning has been widely used to solve combinatorial optimisation problems~\cite{bello2016neural}~\cite{deudon2018learning}~\cite{kool2018attention}~\cite{nazari2018reinforcement}, especially in routing problems.  Deep reinforcement learning can be further divided into construction and improvement reinforcement learning methods. Construction methods construct a solution one node (for routing problem) at a time~\cite{bello2016neural}. Improvement type deep reinforcement learning methods combine machine learning with existing local improvement heuristic methods, where an initial feasible solution is iteratively improved upon to find a better solution. Such improvement based reinforcement learning methods generate state of the art results~\cite{wu2016training}~\cite{da2020learning} for TSP and ~\cite{hottung2019neural}~\cite{chen2019learning}~\cite{lu2019learning} CVRP. We briefly summarise the work below.
%This work focuses on optimising route problems using policy gradient methods. Due to their colossal search space, route (combinatorial) problems are difficult to model and hard to generate episodes and experiences of all probable states and actions (policy gradient methods rely on using a neural network to learn the state-action pairs). 

Bello et al.~\cite{bello2016neural} was one of the first to propose a reinforcement learning approach to solving Combinatorial optimisation problems. In their approach, they employed Pointer Network proposed by Vinyals et al.~\cite{vinyals2015pointer} and trained pointer network with the actor-critic reinforcement learning approach. They proposed neural reinforcement learning to optimise policy using the policy gradient methods. They demonstrated its use for TSP and knapsack problems. Nazari et al.~\cite{nazari2018reinforcement} developed a model based on pointer network~\cite{vinyals2015pointer} to solve CVRP, where the authors use element-wise projections instead of LSTM in the pointer network encoder. They also apply this model to VRP with split deliveries and a stochastic variant.

Deudon et al.~\cite{deudon2018learning} proposed a neural framework for TSP based on the self-attention network proposed by Vaswani et al.,~\cite{vaswani2017attention}. The authors used an attentive encoder as input to encode the cities, and the decoder sequentially generates the partial tour. The network is trained via actor-critic reinforcement learning, and the solution is improved with the traditional 2-Opt search method. Another model called, Attention Model(AM)~\cite{kool2018attention} also uses a standard transformer to encode the cities. The decoder is decoding sequentially with a query composed of the first and the last city in the partial tour. The decoder also decodes a global representation of all cities. They trained the model with reinforcement and a deterministic baseline (use a rollout baseline). The attention model used a greedy rollout baseline based on REINFORCE to train the network. Attention Model(AM)~\cite{kool2018attention} has been applied to TSP and routing problems, including VRP and orienteering problems(OP). Peng et al.~\cite{peng2019deep} show that using the attention model dynamically can enhance its performance.%Peng et al.~\cite{peng2019deep} show that using the Attention Model dynamically can enhance its performance.

A recent work augments a graph network with Monte Carlo Tree Search (MCTS) to improve the search exploration of tours by evaluating multiple following nodes (cities) in the tour. They use a graph neural network with Monte Carlo Tree Search (MCTS)~\cite{xing2020graph} which is based on AlphaGo~\cite{silver2016mastering}. They show that this technique ameliorates the search exploration, which cannot go back once the selection of the nodes is made.

Recent improvement method, LHI~\cite{wu2016training} propose to learn the heuristics for routing problems based on the deep reinforcement learning framework. As a policy network, they design a self-attention based deep architecture. LHI has been applied to TSP, CVRP. Another local search heuristic method, based on 2-Opt heuristics proposed by Da et al.~\cite{da2020learning}. The author has designed a transformer-based network for learning by reinforcement learning and actor-critic to select nodes. They also introduce a policy neural network that leverages a pointing attention mechanism, easily extended to more general k-opt moves. 

POMO~\cite{kwon2020pomo} identify symmetries in reinforcement learning methods for solving CO problems that lead to multiple optima. Such symmetries can be leveraged during neural net training via multiple parallel rollouts, each trajectory having a different optimal solution as its goal for exploration. POMO solved three NP-hard problems aforementioned, namely TSP, CVRP, and KP, using the same neural net and the same training method. ERRL~\cite{sultana2021learning} method incorporates an entropy term, defined over the policy network's outputs, into the loss function of the policy network. Hence, policy exploration can be explicitly advocated, subjected to a balance to maximise the reward. As a result, the risk of pre-convergence to inferior policies can be reduced. ERRL has been applied to TSP, CVRP and MRPFF. EPOSE belongs to the category of construction based reinforcement learning method, which is an extension of ERRL~\cite{sultana2021learning} method. However, in contrast to ERRL~\cite{sultana2021learning} methods, for sample efficiency, we propose off-policy learning (discussed in Section~\ref{sec:epose}). Also, ERRL~\cite{sultana2021learning} encourages exploration using hyperparameter tuning manually; instead, we automatically adjusted hyper-parameters to encourage exploration.

\section{Background}\label{sec:BG}

In this section, we first describe the notation used and maximum entropy reinforcement learning model. 

\subsection{Notation}
Table~\ref{tabile:notation} shows the notation we use in this paper.

\begin{table*}[!b]
\small
\centering
\caption{Summary of symbols.}
%The gap percentages reported with respect to optimal value for CVRP100 15.67 using LKH3 \cite{helsgaun2017extension}
\begin{tabular}{|l|l|}\hline
%Parameter(Learning Rate) & Parameter(Co-efficient)    &  TourL       & Gap \\ \hline
%Co-efficient &0.5&0.6&0.7&0.8&0.9\\\hline
\textbf{Symbols} & \textbf{Definition}\\\hline
% 0.00001 &&   &  & &&\\ \hline
% 0.0001  & &&&   & & \\\hline
%G & Input graph  \\ \hline
v & Each node.\\ \hline
%$m_i$ & Coordinates of node i\\ \hline
$I$& A problem instance.\\ \hline
$\pi$& Solution.  \\\hline
$L(\pi|I)$  &Length of the tour defined for permutation $\pi$ given problem instance $I$.  \\ \hline
%$s = \pi_{1:t-1}$ & State (partial solution)\\\hline
$s$& State represented as a partial solution or a sequence of previously selected
actions, $a_{1:t-1}$. \\\hline
$a$ & Action is defined as selecting one of the unvisited nodes.\\ \hline
$a_{1:t-1}$  & A sequence of previously selected
actions.\\ \hline 
$p_{\phi}(\pi|I)$ & Stochastic policy based on the attention (encoder-decoder) model that generates solution $\pi$ from given problem instance $I$.\\ \hline
%$\pi_\phi(.|s_t)$ & Denote the policy which means distribution over all actions given state \\ \hline
%H(\pi_\phi(.|s_t)$ & Entropy of the policy \\ \hline
%$\tau_\pi$  & Distribution over all possible actions in a given state induce by policy \\\hline
$\alpha$& Entropy parameter.\\\hline
%$\lambda$  &Discount rate \\ \hline
%$V(s)$  &Value function (state value function) \\ \hline
Q function  & The expected reward for an action taken in a given state following policy.\\ \hline
% $r(s_t, a_{t})$ & Immediate reward \\ \hline 
$R(\pi^i)$ & Reward. \\ \hline 
$\tau_{\pi|I}$  & Distribution of sequence in terms of state, action and reward following stochastic policy $p(\pi|I)$. \\ \hline
D  & Replay buffer. \\ \hline
%$V_\theta\textbf{'}{(s_{t+1})}))$  &Target network for Q function estimation \\ \hline
$\eta$& Target smoothing coefficient.\\\hline
%$H^{'}$ & Target Entropy  \\ \hline
%$Z^{\pi}$  &Partition function \\ \hline
$V_\theta\textbf{`}{(s_{t+1})}]$ &Target network estimation for Q, next state $s_{t+1}$. \\ \hline
${E}_{s_{t+1}}$& Expectation of state-value of next state.\\\hline
\end{tabular}
\label{tabile:notation}
\end{table*}

\subsection{Maximum entropy Reinforcement Learning}
In standard reinforcement, the learning objective is the expected sum of rewards, and the goal is to learn a policy that maximises rewards~\cite{sutton2011reinforcement}. Sultana et al.~\cite{sultana2021elearning} proposed a method that augments the reward with an entropy term~\cite{ahmed2018understanding}. They proposed an on-policy based method, where the learning evaluates and improves the same policy the agent is already using for action selection. Entropy is directly related to the unpredictability of an agent's actions in a given policy. Entropy prevents policies from becoming deterministic too quickly, and the use of entropy is to encourage exploration~\cite{sultana2021elearning}. The entropy here ensures that the model is more likely to take another action to encourage exploration. They calculate the entropy of the policy that supports help with exploration by encouraging the selection of more stochastic policies, used hyper-parameter values to control the trade-off between optimising for the reward and the entropy of the policy. However, they set the hyper-parameter value manually. This parameter needs to tune for each routing problem in the previous work~\cite{sultana2021learning}. Using fixed values manually is a poor solution since the policy should be free to explore more in regions where the optimal action is not certain but remain more deterministic in states with a complete distinction between good and bad actions. Therefore, we propose off-policy based learning, called EPOSE. The off-policy learning algorithm is based on behaviour and estimation policies. The behaviour policy is used to select an action (unrelated to the evaluated policy). The evaluated policy learns the value function for the policy improvement~\cite{sutton2011reinforcement}. Instead of only seeking to maximise the lifetime rewards, EPOSE seeks to maximise the entropy of the policy, adjusting the parameters automatically using gradient descent. We use an \textit{automatic gradient-based} temperature tuning method that adjusts the expected entropy over the visited states, and we find it largely eliminates the need for hyper-parameter tuning.

%The magnitude of the reward differs across problems and depends on the policy, which improves over time during training. The optimal entropy depends on this magnitude; therefore, adjusting the temperature is difficult. The entropy can vary unpredictably across tasks and during training. Hence, 

% \subsection{Q function}\label{sec:Q} 
% Q function calculates the quality of a state–action combination $Q: s \times a $. At each time t when the agent selects an action a, observes an immediate reward r(a,s), enters a new state $s_{t+1}$ which may depend on both the previous state and the selected action and Q is updated. As previously mentioned, to increase sample efficiency here, we use two Q-functions which means we have previous samples (behaviour policy), also we have a target network to estimate Q, which is the estimation policy. Q function shows a measure of expected reward, Q value provides how good an action is given a state following policy. In this work, we maintain Q networks that we optimise using Equation~\ref{equation:bell}.
\section{Problem Formulation}

We demonstrate the effectiveness of EPOSE on three typical routing problems, namely the TSP, CVRP and SDVRP. To avoid excessive details, we focus on explaining the EPOSE approach in terms of the TSP. The approach works similarly for the other problems (CVRP and SDVRP). We described the EPOSE in terms of CVRP and SDVRP in the Appendix. We are given a TSP instance with a group of nodes $\{v_1,\cdots,v_m\}$. Here m is the number of nodes. Our goal is to find a solution $\pi$ given a problem instance $I$ so that each node can be visited exactly once and the total tour length is minimised. The length of a tour is defined for permutation $\pi$ as :%Then, Then, the solution of TSP can be represented as the permutation $\pi$ of input sequences:The TSP graph can be represented as a sequence of nodes $\{v_1,\cdots,v_m\}$ in 2D Euclidean space. nd $x_i$ denotes the coordinates of node i. 
%$I= x_{i=1}^{N}$ where $x_i\in R^2$. 

\begin{equation}
\begin{aligned}
L(\pi|I)=   {\parallel x_{\pi_m} -  x_{\pi_1} \parallel}_{2} + \sum_{t=1}^{m-1} {\parallel x_{\pi_t} -  x_{\pi_{t+1}} \parallel}_{2}
\label{equation:ttt}
\end{aligned} 
\end{equation}

Where ${\parallel \cdot \parallel}_2$ denotes the L2 norm.

Our learning problem can be defined as a policy search in a markov decision process, i.e. for every state, you have probability distribution of actions to take from that state. The reinforcement learning is defined by the tuple ($s$, $a$, $T$, $R$), where State and action is assumed to be discrete.~\textbf{State} $s$ is represented as a sequence of previously selected actions: $a_{1:t-1}$.~\textbf{Action} $a$ is defined as selecting one of the unvisited nodes; $T$ is the deterministic transaction function ($T: s \times a \rightarrow s)$, $R$ is the reward function ($R: s \times a \rightarrow \mathcal{R})$. 
\begin{figure*}[!b]
\begin{center}
\includegraphics[width=0.7\textwidth]{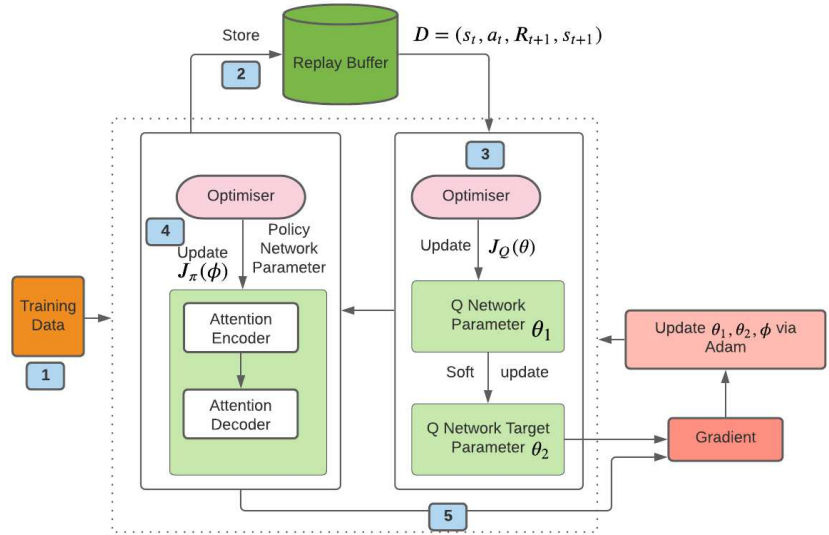}
\caption{Overview of EPOSE. 1: Training data. 2: We store all experiences gathered from the environment ($s_t,a_t,R_{t+1},s_{t+1}$) 3: To increase the sample efficiency, we use two Q-functions, which means we have previous samples (behaviour policy), and we have a target network (a target network estimation for Q, which is the estimation policy). 4: We use the Q function calculated in the policy improvement step to guide change in the policy 5: We learn the policy using the estimation of the gradient of the expected return to the policy parameters.}\label{fig:epose}
\end{center}
\end{figure*}

\section{Exploration based Policy Optimisation with Sample Efficiency (EPOSE) Model}\label{sec:epose}

The off-policy based method stores all experiences in the replay buffer gathered from the environment, denoted as $D$. Essentially, the off-policy-based method uses experience replay, i.e. randomly draws the sample from the replay buffer. Off-policy updates policy differs from the behaviour policy, which means the off-policy estimates the reward for future actions and appends a value to the new state without following a greedy policy~\cite{sutton2011reinforcement}. However, the on-policy reinforcement learning algorithm estimates the value of the policy, and the policy performance depends on the last rewards and updates the policy based on the reward~\cite{sutton2011reinforcement}. Essentially, the on-policy updates the parameters that predict the value of a specific state/action~\cite{sultana2021learning}. The off-policy updates the policy based on stored in experience replay, evaluating actions from a policy that is no longer current. Experience is modelled as $s_t,a_t,R_{t+1},s_{t+1}$, which are sampled from the replay buffer to update policy, i.e., the observed state of the environment, $s_t$, takes action $a_t$ based on the policy. Then, the agent gets a reward $R_{t+1}$ and next state $s_{t+1}$. 

Also, EPOSE optimises Q-function\footnote{The Q function calculates the quality of a state–action combination, $Q: s \times a $. At each time $t$ when the agent selects  action a, observes an immediate reward and enters a new state $s_{t+1}$ which may depend on both the previous state and the selected action and Q is updated.} to increase the sample efficiency. The Q function shows a measure of expected reward, and the Q value determines how good an action is given a state following the policy. To optimise the Q function, EPOSE uses two networks. One is for the previous samples, and another is for estimating Q. One network is used to select the action (a.k.a., the behaviour policy), and the other one used to evaluate the policy, i.e. a target network to estimate Q (a.k.a. the estimation policy). The EPOSE algorithm is summarised in Figure~\ref{fig:epose}. 

%In this work, we maintain Q networks that we optimise using Equation~\ref{equation:bell}.

EPOSE consists of the following steps:

First, following Figure~\ref{fig:epose}, in steps 1 and 2, using training data, the method optimises Q function in every update step using the gradient of the mean square loss between the predicted action value and the target action value to improve the sample efficiency and training speed:

\begin{equation}
\begin{aligned}
J_{Q}(\theta) =  {E}_{({s_{t},a_{t})\sim D}}[\frac{1}{2} (Q_\theta(s_t,a_{t}) - q_t)^2].
\label{equation:bell}
\end{aligned} 
\end{equation} 

where $q_t =  (r(s_t, a_{t})+ {E}_{{s_{t+1}\sim \tau(\pi_t)}  [V_\theta\textbf{`}(s_{t+1})}])$,

Here, $r(s_t, a_{t})$ is the immediate reward plus the expected value of the next state, $V_\theta\textbf{`}(s_{t+1})$, i.e. for the state-action pairs in the experience replay buffer, we are minimising the square difference between the prediction of our Q function and the immediate reward plus the expected value of the next state, which is called the target value. In this step the Q values $Q_\theta(s,a)$ trained from replay memory D on $s_t,a_t,R_{t+1},s_{t+1}$ transitions by minimising the mean square loss (Equation~\ref{equation:bell}).

Second, in the policy optimisation step, we use optimised Q values. Following Figure~\ref{fig:epose}, in step 4, after we draw N sample problems, we compute reward $R(\pi^i)$ for each solution $\pi^i$. To maximise the expected return, we use gradient descent in step 5 with an approximation:

\begin{equation}
\begin{aligned}
J_{\pi \sim p_\phi(\pi|I)}(\phi) = (R(\pi) -\vartheta_\phi(s)) logp_\phi(\pi|I) +\alpha  H (({\pi|I})_{\phi} (. |s))\label{equation:mc},
\end{aligned} 
\end{equation} 

where $p_\phi(\pi^i|I) =\prod_{t=1}^{N} p_\phi(a^i{_t}|s, a_{1:{t-1}})$. 
%In step 4 and 5, we optimise cost $L(\pi|I)$ (tour length) by gradient descent, gradient estimator with critic $\vartheta_{\phi}(s)$:The critic network is considered as the baseline and output a scalar $\vartheta_\phi(s)$ to estimate the cumulative rewards, i.e., for solution (a sequence of nodes) from problem instance $I$ as tour length: $ L(\pi|I)$. 

Here, $a_{1:t-1}$ is a sequence of previously selected actions; $R(\pi)$ is the reward (return) for each solution $\pi$; and H is the entropy of the policy and $\vartheta_\phi(s)$ to estimate the cumulative rewards, i.e. for the solution (a sequence of nodes) from problem instance $I$. %In step 4 and 5, we optimise cost $L(\pi|I)$ (tour length) by gradient descent, gradient estimator with critic $\vartheta_{\phi}(s)$: The critic network is considered as the baseline and output a scalar $\vartheta_\phi(s)$ to estimate the cumulative rewards, i.e., for solution (a sequence of nodes) from problem instance $I$ as tour length: $ L(\pi|I)$. 

To the best of our knowledge, we are the first to use an actor-critic off-policy gradient-based method to optimise route problems. In Figure~\ref{fig:epose}, our actor network refers to the graph attention model (step 4) described in Appendix. The critic network comprises multiple one-dimensional convolutional layers and shares the same encoder network with the actor. The critic network estimates the reward for a state, which uses three attention layers, similar to our encoder. The node embeddings are averaged and processed by a multi-layer perceptron with one hidden layer with 128 neurons and ReLu activation and a single output. The critic network outputs $\vartheta_\phi(s)$ to estimate the cumulative rewards, i.e., length of the tour~$L(\pi|I)$. Here, $\alpha$ represents the entropy parameters, i.e. the randomness of the policy versus the reward.

At the end, in addition to the Q function and the policy, we also learn $\alpha$, the temperature parameter, automatically. The aim is to find a stochastic policy with the maximum expected return that satisfies a minimum expected entropy constraint. Using the entropy term EPOSE seeks to maximise the entropy of the policy to encourage exploration, which helps the policy to assign equal probabilities to actions that have the same or nearly equal Q values. We formulate a maximum entropy learning objective that improves the reward over time during training. We learn $\alpha$ by minimising the dual objective (approximating dual gradient descent~\cite{boyd2004convex}) in Equation~\ref{equation:alpha}.

\begin{equation}
\begin{aligned}
{\alpha}_t =  argmin_{\alpha_t} {E_{a_t \sim \pi_t}}[-\alpha_t(log \pi_t(a_t|s_t; \alpha_t) - \alpha_t H)]
\label{equation:alpha}
\end{aligned} 
\end{equation} 
%therefore, we change our calculation of the temperature parameter loss to reduce the variance of the estimation and let the policy explore more. 

We assign the optimised $\alpha$ value in Equation~\ref{equation:mc} to optimise the policy, because forcing the entropy as a fixed value~\cite{sultana2021learning} would be inferior solution since the policy needs to be free to explore. Therefore, using Equation~\ref{equation:hcon}, we automatically adjust the entropy over the state to compute gradients for $\alpha$ with the following objective (Algorithm~\ref{algo:ERRam}, step 16), and this step eliminates the need to tune the hyper-parameters:

\begin{equation}
\begin{aligned}
J(\alpha) =  E_{a_t \sim \pi_t}[-\alpha log \pi_t (a_t|s_t) -\alpha H)]
\label{equation:hcon}
\end{aligned} 
\end{equation} 

Overall, this work optimises objective functions ($J_Q(\theta)$), $J_\pi(\phi)$ and $J(\alpha)$. This work first optimises the Q function. The Q function calculates the quality of a state–action combination $Q: s \times a $. At each time $t$ when the agent selects action a, observes an immediate reward, enters a new state $s_{t+1}$ which may depend on both the previous state and the selected action and Q is updated. We use two Q functions to increase the sample efficiency. In this work, we maintain Q networks that we optimise using Equation~\ref{equation:bell}. After optimising the Q function using Equation~\ref{equation:mc}, we maximise the reward, which is an optimised tour length. Overall our model first optimises the Q function, which has a greater effect on the sample efficiency, and Q function optimisation helps the model to improve the policy that maximises the expected reward. The final objective is to optimise the temperature loss, which helps to reduce variance in the estimation. In particular, we parameterise two Q functions, with parameters ($\theta_i$), and train them independently to optimise $J_Q(\theta_i)$ (Algorithm~\ref{algo:ERRam} step 17). Instead of setting the parameters manually, we automate the process in this work. Algorithm~\ref{algo:ERRam} summarise the EPOSE method. 
%In table~\ref{tabile:notation}, we notified the action and state for our problems. 

Our model uses encoder-decoder architecture. The encoder attention layers contain multi-head attention layers and key, value and query dimensions. The feed-forward sub-layer in each attention layer has a dimension of 512. The decoder generates sequence $\pi$. The decoder takes the graph embedding and node embedding, a problem-specific mask and a single-context node embedding as input. The context consists of the graph embedding and the embedding of the first and last nodes (previously output) of the partial tour at each time step. Nodes that have already been visited are masked. The input of the decoder uses a single context node embedding. For the TSP, when a partial tour has been constructed, it cannot be changed, and the remaining problem is to look for a path from the last node to the first node through all unvisited nodes. The model is the same for other problems but only needs to change the input, masks and decoder context vectors accordingly following attention model~\cite{kool2018attention}. All of our EPOSE experiments use the Attention Model (which we refer to as the "original Kool et al.~\cite{kool2018attention}"), whose details are given in Appendix.

\begin{algorithm}
\caption{EPOSE for routing problems}\label{algo:ERRam}
%\LinesNumbered
\footnotesize
\SetKwFunction{This}{this}
Input: number of epochs E, steps per epoch T, batch size B,

Initialise $Q_{\theta1} :$ $s\leftarrow r ^{\mid a \mid}$; $Q_{\theta2}:$ $s\leftarrow r ^{\mid a \mid}$ ; $\pi_\phi : s\leftarrow [0,1]^{\mid a \mid}$ 

Initialise $Q^{'}_{\theta1} :$ $s\leftarrow r ^{2\mid a \mid}$; $Q^{'}_{\theta2}:$ $s\leftarrow r ^{\mid a \mid}$ ;

\For{epoch $1$ $\cdots $ $E$}{
\For{step $1$ $\cdots $ $T$}{
$\theta^{'}_1$ $\leftarrow \theta_1$ $\theta^{'}_2 \leftarrow \theta_{2}$

$ D \leftarrow 0$

 $I_i \sim $ randomInstance() $\forall_i \in$ \{$1$ $\cdots$  $B$\}

         $\pi_i \sim $ sampling ($p_\phi(\pi|I)$)  $\forall_i \in$ \{$1$ $\cdots$ $B$\}

         $s_{i+1} \sim $ $p(s_{i+1}|s_i, a_i)$
         
         Compute $L(\pi|I)$ via $\pi_i$ $I_i$
         
         Put \{${s_i, a_i, L(\pi|I), s_{i+1}}$\} into D
         
        % policy improvement step using KL-divergence
        
        % $ \pi_n = {argmin}D_{KL}\bigg{(}\pi(.|s_t)\parallel \frac{exp(\frac{1}{\alpha}Q^{\pi_{old}}(s_t,.))}{Z^{\pi_{old}}(s_t)}\bigg{)}$
         
for each gradient step do

         $\theta_i \leftarrow$ $\dfrac{1}{B}$ [$\theta_i -  \hat \nabla_{\theta_i} j(\theta_i)$ ] for  $i \in \{1, 2\}$
          
        %   $\phi \leftarrow$ $\dfrac{1}{B}$ [$(L(\pi|I_i) -\vartheta_{\phi}(s_i))  \hat \nabla_\phi logp_\phi(\pi|I_i) +\alpha  H (\pi_{\phi} (. |s_{i}))$] 
        
        $\phi \leftarrow$ $\dfrac{1}{B}$ [$(R(L^i) -\vartheta_\phi(s)) logp_\phi(a^i|s) +\alpha  H (\pi_{\phi} (. |s)$] 
         
          $\alpha \leftarrow$ $\dfrac{1}{B}$ [$\alpha -  \hat \nabla_\alpha j(\alpha)$  ]
          
          $Q_i \leftarrow$  $\eta Q_i + (1 - \eta) \hat Q_i j(\theta_i)$  for $i \in \{1, 2\}$
         
}
return optimised $\theta_1, \theta_2, \phi$
}
\end{algorithm}

\section{{Computational Experiments}}\label{sec:exp} 
In this section, we discuss the experimental setting for the evaluation of the proposed method. First, the model was evaluated using TSP, CVRP and SDVRP instances. Second, we analysed the learning trends. Moreover, we performed an ablation study to directly compare our method with various methods and determine the impact of entropy parameters. Our experiments were designed to investigate the following evaluation procedures: 
 \begin{itemize}
 
    \item The performance of our method was evaluated on the randomly generated TSP instance is shown in Table~\ref{tabile:tsp}.
    
     \item The performance of our method was evaluated on the randomly generated CVRP instance is shown in Table~\ref{tabile:randomdata}.
      \item The performance of our method was evaluated on the randomly generated SDVRP instance is shown in Table~\ref{tabile:sdvrp}.
   \item Compare the convergence of our approach with other models in Figure~\ref{fig:smalltsp}.
     \item We analyse the sample efficiency in Figure~\ref{fig:search}. 
   
\end{itemize} 

The following subsections describe the datasets, network settings, decoding strategy, evaluation criteria, and results of the proposed method.

\subsection{Datasets}\label{section:data}
To evaluate our model, we followed existing works~\cite{vinyals2015pointer}~\cite{bello2016neural}~\cite{kool2018attention}~\cite{deudon2018learning}~\cite{xing2020graph}~\cite{khalil2017learning}~\cite{joshi2019efficient}~\cite{sultana2021learning}~\cite{bresson2021transformer}~\cite{kool2021deep}~\cite{sultana2021elearning} for TSPs evaluated with the same types and sizes of datasets to generate instances with 20, 50 and 100 nodes (cities), using 2D Euclidean distance to calculate the distance between two cities, and the objective was to minimise the total travel distance. The coordinates of the city locations were sampled from a uniform distribution ranging from 0 to 1 for both dimensions independently. The dataset generated from the unit square $[0, 1]\times[0, 1]$ and 1,000 test instances was generated with the same data distribution as existing studies. For the CVRP, the vehicle capacities were fixed as 30, 40 and 50 for problems with 20, 50 and 100 nodes (cities), respectively, and the demands of each depot city were sampled from integers $1 \cdots 9$. The SDVRP is a generalisation of the CVRP in which every node can be visited multiple times, and only a subset of the demand has to be delivered at each visit (more settings introduced in the Appendix). 

%Our framework can be applied to many routing problems. 

\subsection{Network Setting}

In this section, we conducted experiments on three routing problems to verify the effectiveness of our method. Among them, the TSP and CVRP are the most widely studied. We trained the proposed attention model for TSP and CVRP instances with nodes n = 20, 50 and 100 (trained 100 epochs). As the testing process did update the model parameters, a larger batch size could be used. We used Intel Xeon 2.4 GHz with 56 cores to complete the training of routing problems. The values of the hyper-parameters used for the training process are listed in Table~\ref{tabile:hyper}. The hyper-parameter values for all problems of the same size were identical. The model was constructed using the PyTorch~\cite{paszke2017automatic} framework and implemented using Python 3.7. The transformer encoder had three layers with 128 dimensional features and eight attention heads. In every epoch, we processed, 2500 batches of 512 instances (although we used $2,500 \times 256$ for the CVRP to fit the GPU memory constraint). We used the Adam optimiser~\cite{kingma2014adam} with a learning rate of $0.0001$ for the optimisation. 

For our attention model, the node embedding was 128 dimensional. The encoder had six attention layers, where each layer contained multi-head attention layers with head number M = 8 and the dimensions of the key, value and query dimensions. The feed-forward sub-layer in each attention layer had a dimension of 512. This set of hyper-parameters was also used for the CVRP and SDVRP.

\subsection{Decoding Strategy}

Few previous studies on learning route optimisation problems included search strategies, such as beam search, neighbourhood search and tree search. Bello et al.~\cite{bello2016neural} proposed search strategies such as sampling and active search. We used the following two decoding strategies:

\textbf{Greedy decoding:} 
Generally, a greedy algorithm selects an optimal local solution and approximates the solution. In each decoding step, the model selects the node with the highest probability in a greedy manner, and all visited nodes are masked. For the TSP problem, the search is terminated when all nodes have been visited. For the CVRP and SDVRP, the search process ends when the requirements of all nodes are satisfied to construct an effective solution.

\textbf{Sampling:} 
In each decoding timestep, the random policy samples the nodes, and the nodes are selected according to the probability distribution to construct a solution. Stochastic sampling is usually needed to explore the environment to obtain a better model performance in the training process. In the testing process, we sampled 1,280 solutions following existing studies~\cite{kool2018attention}~\cite{bello2016neural} using stochastic sampling.

\subsection{Evaluation} 
We report the following metrics to evaluate the performance of our model following existing studies~\cite{kool2018attention}~\cite{kwon2020pomo}~\cite{sultana2021learning}:
%over 1,000 test instances
\begin{itemize}
 \item\textbf{Predicted tour length:} The average predicted tour length.
 \item \textbf{Optimality gap:} 
The average percentage ratio of the predicted tour length relative to optimal solutions \cite{joshi2019efficient}.
 \item\textbf{Time:} 
This refers to the total running time in minutes (m) and seconds (s) of 1,000 instances.
\end{itemize}

\begin{table*}[h]
\small
\centering
\caption{The values of the hyper-parameters used in our model.}
%The gap percentages reported with respect to optimal value for CVRP100 15.67 using LKH3 \cite{helsgaun2017extension}
\begin{tabular}{|l|l|}
\hline
%Parameter(Learning Rate) & Parameter(Co-efficient)    &  TourL       & Gap \\ \hline
%Co-efficient &0.5&0.6&0.7&0.8&0.9\\\hline
Hyper-parameter&Value\\\hline
% 0.00001 &&   &  & &&\\ \hline
% 0.0001  & &&&   & & \\\hline
Replay buffer size  &1,000,000  \\ \hline
Fixed entropy parameter&0.03 \\ \hline
Target smoothing coefficient &0.005 \\ \hline
Optimiser  &Adam \\ \hline
Loss &Mean squared error\\ \hline
Entropy target   & $0.98 * (-log (1 / |A|))$\\\hline
\end{tabular}
\label{tabile:hyper}
\end{table*}
\begin{table*}[h]
\centering
\caption{Performance of the methods with respect to Concorde for the TSP (lower is better, bold is best). The running times are reported in seconds (s). In the Type column: H: heuristics, M: meta-heuristics, SL: supervised learning, RL: reinforcement learning, S: sample search, G: greedy search. $-$ denotes the never-tested running time using their codes. \textbf{Lower is better, best in bold}.}
\scriptsize
%\begin{tabular}{lc|cc|cc|cc}\hline
%\multicolumn{2}{c}{} & \multicolumn{2}{c}{TSP=20} & \multicolumn{2}{c}{TSP=50} & \multicolumn{2}{c}{TSP=100} \\ \hline
%\multicolumn{2}{c|}{Method} & Obj & Gap(\%) & Obj & Gap(\%) & Obj & Gap(\%)  \\\hline\hline 
% \begin{tabular}{|l|ll|ll|ll|}
\begin{tabular}{|ll|lll|lll|lll|}\hline
\hline
\multicolumn{2}{|c|}{Method}& \multicolumn{3}{|c|}{TSP20} & \multicolumn{3}{|c|}{TSP50}& \multicolumn{3}{|c|}{TSP100} \\ \hline
\textbf{Method} & Type   & TourL  & Gap(\%)& Time & TourL  &  Gap(\%) & Time & TourL   & Gap(\%) & Time\\\hline
Concorde  & Solver    & 3.83  & 0.00&  4(m) & 5.70  & 0.00& 10(m) & 7.77  & 0.00&55(m)\\\hline
\textbf{Non-learning baselines} &  &&  & &  &  &  &&&  \\\hline
LKH3 & Heuristics& 3.83  & 0.00 & 18(s) & 5.70 & 0.00 &5(m)& 7.77 &0.00&21(m)\\ 
Or-tools& Meta-heuristics &  3.85 & 0.52& -& 5.80  & 1.75 &-& 8.30  & 6.82& -\\\hline
\textbf{Learning Models}& &  & &&& & &  &  &    \\\hline
Sultana et al.~\cite{sultana2022learning}& SL& 3.85 &  0.526&- & 5.85  & 2.63 &-& 8.31 &6.94&-\\
Bello et al.~\cite{bello2016neural} & RL, G& 3.89 &  1.56&- & 5.99  & 5.08 &-& 9.68  &24.73&-\\
Kool et al.~\cite{kool2018attention}& RL, G& 3.85 &0.52 & 0.1(s)& {5.80}&1.75& 2(s)  & {8.15}  &4.89& 6(s)\\
Kool et al.~\cite{kool2018attention}& RL,S & 3.84  & 0.26&5(m) & 5.75 & 0.87& 19(m)& 7.94 &2.18&55(m)\\
ERRL.~\cite{sultana2021learning}  & RL,G &3.83& *& 0.1(s) &{5.73}  & 0.52& 2(s) & {7.85}  & 1.02 & 6(s)\\
POMO \cite{kwon2020pomo}&RL & {3.90} & 1.82 & 0.1(s) & {5.73}  & 0.52 & 1(s)& {7.85} & 1.02 & 2(s)\\\hline
\textbf{EPOSE} & RL,G&3.81 & * & 0.1(s) & {5.70} & *& 2(s)& 7.77 &* & 6(s)\\
\textbf{EPOSE} & RL,S&3.78& * & 1(m) & {5.64} & *& 17(m)& 7.74 &* & 49(m)\\\hline
%\textbf{[EPOSE[EBRL]}& RL,G&3.81 &* & 0.1(s) & {5.70} & *& 2(s)& 7.77 & * & 6(s)\\\hline
\end{tabular}
\label{tabile:tsp}
\end{table*}

\subsection{Results}
In  this  section,  we  compare  our  model's experimental results with those of existing methods.

\subsubsection{TSP results}
TSP is defined as finding the shortest tour which visits each of the cities once and returns to the starting city, given the distances between each pair of the cities. For TSP, we report baselines from three different categories in Table~\ref{tabile:tsp}. The first category includes specialised solvers using Concorde~\cite{applegate2006concorde} as an optimal solver. The second category consists of machine learning approaches that construct solutions sequentially and use greedy inference comparing our model with learning approaches by Kool et al.~\cite{kool2018attention}, POMO~\cite{kwon2020pomo}., Sultana et al.,~\cite{sultana2021learning}. The third category also includes learning-based approaches, but using greedy and sampling search inference. In Table~\ref{tabile:tsp} we report our main results for EPOSE and compare the performance of EPOSE on TSP with other baselines. EPOSE with greedy decoding, and sampling outperformed compared to the result generated by Concorde~\cite{applegate2006concorde}. EPOSE also outperformed all other learning-based heuristics significantly. 

We point out that the running time of the greedy decoding strategy is much faster compared to the sampling strategy Table~\ref{tabile:tsp} (shown in the first column indicated by G as greedy and S as sampling). We reported the solution time (from test instances) of some methods from our implementation as it is not worth comparing the running time reported by others because run on the same hardware is necessary. 
%POMO~\cite{kwon2020pomo}, ERRL~\cite{sultana2020learning}.Our model, Kool et al.~\cite{kool2018attention}, use greedy while Nazari et al.~\cite{nazari2018reinforcement} uses beam search.

\begin{table*}[h!]
\centering
\caption{Performance of the methods with respect to LKH3 for the CVRP (lower is better, bold is best). The running times are reported in minutes (m) and seconds (s). In the Type column: H: heuristics, M: meta-heuristics, SL: supervised learning, RL: reinforcement learning, S: sample search, G: greedy search. $-$ denotes the never-tested running time using their codes. \textbf{Lower is better, best in bold}.}
\scriptsize
\begin{tabular}{|ll|lll|lll|lll|}\hline
\multicolumn{2}{|c|}{Method} & \multicolumn{3}{|c|}{CVRP20} & \multicolumn{3}{|c|}{CVRP50}& \multicolumn{3}{|c|}{CVRP100} \\ \hline
\textbf{Solver} &Type   & TourL  & Gap(\%)& Time & TourL  &  Gap(\%) & Time & TourL   & Gap(\%) & Time\\\hline
LKH3 &Type & 6.14  &  0.00 & 120(m)  & 10.39 & 0.00\% & 420(m) &  15.67 &0.00 & 780(m) \\ \hline
\textbf{Non-learning baselines}    &  & &  &   &   &  &    &  & &\\\hline
Or-tools&Type &  6.43& 4.73   &- &11.43 & 10.00  & -& 17.16& 9.50 &-  \\\hline
\textbf{Learning Models} &  &  & &  &  &  &  &  & & \\\hline
Nazari et al.~\cite{nazari2018reinforcement}&RL,G & 7.07& 15.14& -  & 11.95& 15.01 & - &  17.89 &14.16 & - \\
Kool et al.~\cite{kool2018attention}&RL,G & {6.50} & 5.86& 0.1(s)  & 10.89 & 4.81& 1(s)  & {16.99} &8.42&  3(s)\\
Kool et al.~\cite{kool2018attention}&RL,S & {6.35} & 3.42& 4(m)  & 10.60  & 2.02& 25(m)  & {16.23} &3.57&  110(m)\\
{ERRL}~\cite{sultana2021learning} &RL,G  & 6.34 & 3.25 & 0.1(s)  & 10.77 & 3.65 & 1(s)  & {16.35} & 4.33 &  6(s) \\
POMO\cite{kwon2020pomo}&RL & 6.94 & 13.02& 1(s)   & 11.03 & 6.15& 2(s) & {16.35} &4.33&  9(s) \\\hline
\textbf{EPOSE}&RL,G  & {6.24} & 1.62  & 0.1(s) & {10.73} & 3.27 & 1(s) & {16.02} & 2.23 & 6(s)\\
\textbf{EPOSE}&RL,S  & {6.17} & 0.48 & 2(m) & {10.61} & 2.11 & 19(m) & {15.91} & 1.53 & 92(m)\\\hline
%\textbf{EPOSE[EBRL]}& RL,G& {6.24} & 1.62  & 0.1(s) & {10.73} & 2.63 & 1(s) & {16.20} & 3.38 & 3(s) \\\hline
%\textbf{ERRL2} & {6.67} & 8.63  & 5.4  & {11.01} & 2.63 & 14  & {17.23} & 9.95 & 33 \\\hline
\end{tabular}
\label{tabile:randomdata}
\end{table*}

\subsubsection{CVRP Results}
The CVRP generalises the TSP, where the starting city must be a depot and every other city has a demand to be served by the vehicles. Multiple routes can be planned in the CVRP, each for a different vehicle, which visits a subset of customers with total whose demands do not exceed the capacity of the vehicle. All customers need to be covered by the routes. For the CVRP, we also reported baselines. First, we included specialised solvers, where all problems were solved using one run of LKH~\cite{helsgaun2017extension}. Then, the meta-heuristic OR-Tools\cite{perron2019google}. Subsequently, the learning-based approaches that use greedy and sampling search, which include Kool et al.~\cite{kool2018attention}, Nazari et al.~\cite{nazari2018reinforcement}, POMO~\cite{kwon2020pomo}, ERRL~\cite{sultana2021learning}. Table~\ref{tabile:randomm} shows that our approach outperformed all other models that constructed the solution sequentially for all sizes of the CVRP.
\begin{table*}[!ht]
\centering
\caption{Performance of the methods with respect to LKH3 for the Split Delivery VRP. The running times are reported in minutes (m) and seconds (s). In the Type column: H: heuristics, SL: supervised learning, RL: reinforcement learning, S: sample search, G: greedy search. $-$ denotes the never tested running time using their codes. \textbf{Lower is better, best in bold}.}
\scriptsize
%\resizebox{0.4\textwidth}{!}{
\begin{tabular}{|ll|lll|lll|lll|}\hline
%\begin{tabular}{lc|ccc|ccc|ccc}\hline
%\multicolumn{2}{c}{} & \multicolumn{3}{c}{MRPFF=20} & \multicolumn{3}{c}{MRPFF=50} & \multicolumn{3}{c}{MRPFF=100} \\ \hline
%\multicolumn{2}{c|}{Method} & Obj & Gap(\%) & Time(s) & Obj & Gap(\%) & Time(s) & Obj & Gap(\%) & Time(s) \\\hline\hline 
\multicolumn{2}{|c|}{Method} & \multicolumn{3}{|c|}{SDVRP=20} & \multicolumn{3}{|c|}{SDVRP=50}& \multicolumn{3}{|c|}{SDVRP=100} \\ \hline
\textbf{} & & TourL  & Gap(\%) & Time & TourL &  Gap(\%) &  Time& TourL  &   Gap(\%) & Time\\\hline
LKH3  &   & 6.20  & 0.00 & 98(m) &10.35  & 0.00& 336(m) & 15.62 & 0.00 & 660(m)\\ \hline
\textbf{Learning Model}& &  &  & &  &  &  &  &  &  \\\hline
Nazari et al.\cite{nazari2018reinforcement} &RL,G& 6.51 & 5  &- & 11.32 &  9.37& -  & 17.12 &9.60 &  - \\
Kool et al.\cite{kool2018attention} &RL,G& 6.39 &3.06 & 0.1(s) & 10.92 &5.50 &  1(s) & 16.83 & 7.74 & 3(s)  \\
Kool et al.\cite{kool2018attention} &RL,S& 6.25 & 0.806 & 8(m) & 10.59  & 2.31& 40(m)  & 16.27 & 4.16 &  169(m) \\\hline
\textbf{EPOSE} &RL,G&6.18 & * &   0.1(s)  & 10.40  & 0.48 & 1(s) & 16.00 & 2.43  & 3(s) \\
\textbf{EPOSE} &RL,S&6.01&* & 4(m)    &10.27   & *  & 27(m) &15.96  & 2.17  & 140(m) \\\hline
%\textbf{EPOSE[EBRL]}& RL,G&6.11& * & 0.1(s) &10.35  &*& 1(s)& 16.02 & 2.56& 3(s)\\\hline
\end{tabular}%}
\label{tabile:sdvrp}
\end{table*}
% \subsection{MRPFF Results}

% To analyse the EPOSE method, we used MRPFF dataset proposed by Sultana et al~\cite{sultana2021learning} to test our model's performance compared to other existing methods. MRPFF closely related to CVRP, where the starting city must be a depot, and consider the problems setting as CVRP except there is no customer demand, and we consider only one fixed vehicle, visited two sets of customers. For the MRPFF, also we reported baselines from three different categories. MRPFF experiment of applying EPOSE results with 20, 50, and 100 customer nodes are reported in Table~\ref{tabile:mrpff}, and all the EPOSE models outperform all the learning-based methods for MRPFF datasets.

\subsubsection{SDVRP Results}
The SDVRP is a generalisation of the CVRP. It allows every node to be visited multiple times over multiple routes. Very few researchers have reported results, for the SDVRP. For the SDVRP results, we categorised baselines from two different categories. The first category included specialised solvers, where all problems were solved using one run of LKH~\cite{helsgaun2017extension}. The second category consisted of machine learning approaches that use greedy and sampling search, which included Kool et al.~\cite{kool2018attention} and Nazari et al.~\cite{nazari2018reinforcement}. The model by Kool et al.~\cite{kool2018attention} achieved only subpar performance when the problem size increases. We assumed that this had more impact on the SDVRP, as it is a more complex problem than the CVRP. We also wanted to highlight that training our model was much more memory efficient compared to training the standard model by Kool et al.~\cite{kool2018attention}. Our EPOSE outperformed all learning-based approaches that we reported in Table~\ref{tabile:sdvrp}. 

For all problems, in Tables~\ref{tabile:tspp},~\ref{tabile:randomm} and~\ref{tabile:sdvrp} report the solution times based on our implementation from the test instances of the methods. The running time of the greedy search was much faster compared to that of the sampling search.
\begin{figure}
\begin{center}
  \begin{subfigure}[b]{0.45\columnwidth}
    \includegraphics[width=\linewidth]{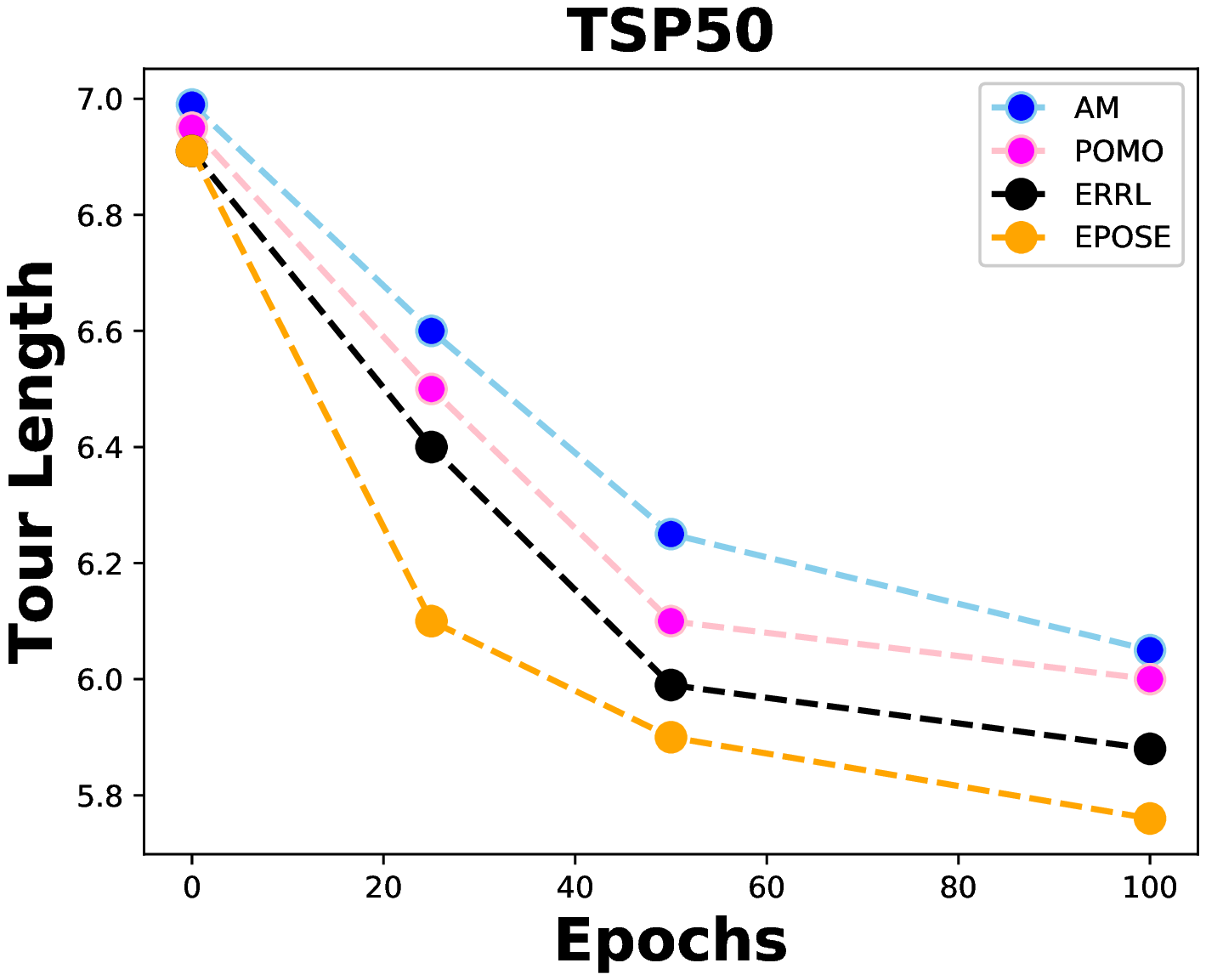}
    \caption{TSP50 Convergence}\label{fig:tsmalltime}
  \end{subfigure}
  \hfill %%
  \begin{subfigure}[b]{0.45\columnwidth}
    \includegraphics[width=\linewidth]{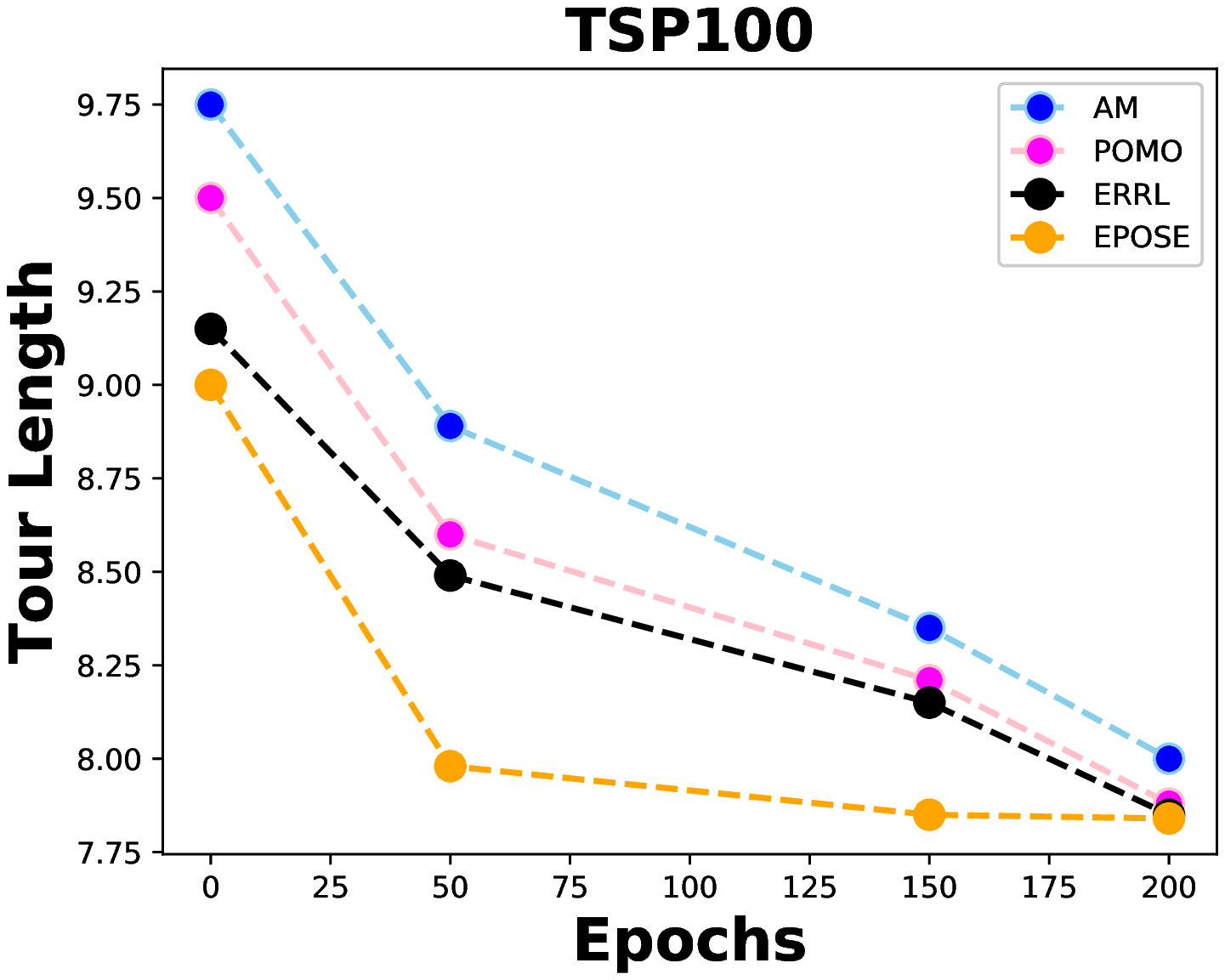}
    \caption{TSP100 convergence}\label{fig:vsmalltime}
  \end{subfigure}
   \caption{The convergence curves of the EPOSE model for TSP50 and TSP100 compared to Kool et al~\cite{kool2018attention} (AM), POMO~\cite{kwon2020pomo} and ERRL~\cite{sultana2021learning}.}\label{fig:smalltsp}
\end{center}
\end{figure}

\subsubsection{Learning Curves}

\begin{figure}
\begin{center}
  \begin{subfigure}[b]{0.33\columnwidth}
    \includegraphics[width=\linewidth]{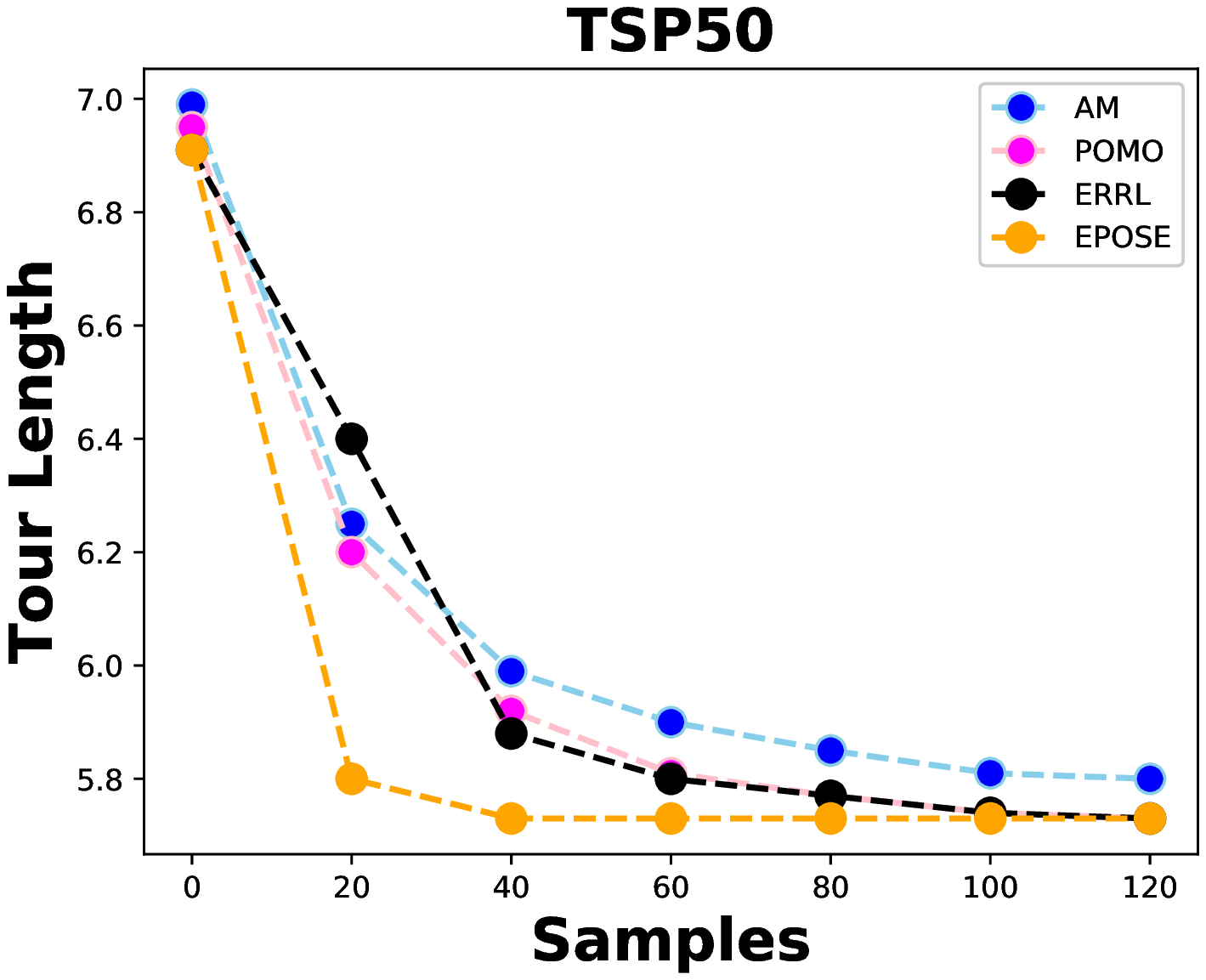}
    \caption{TSP50 sample efficiency}\label{fig:sam}
  \end{subfigure}
  \hfill %%
  \begin{subfigure}[b]{0.33\columnwidth}
    \includegraphics[width=\linewidth]{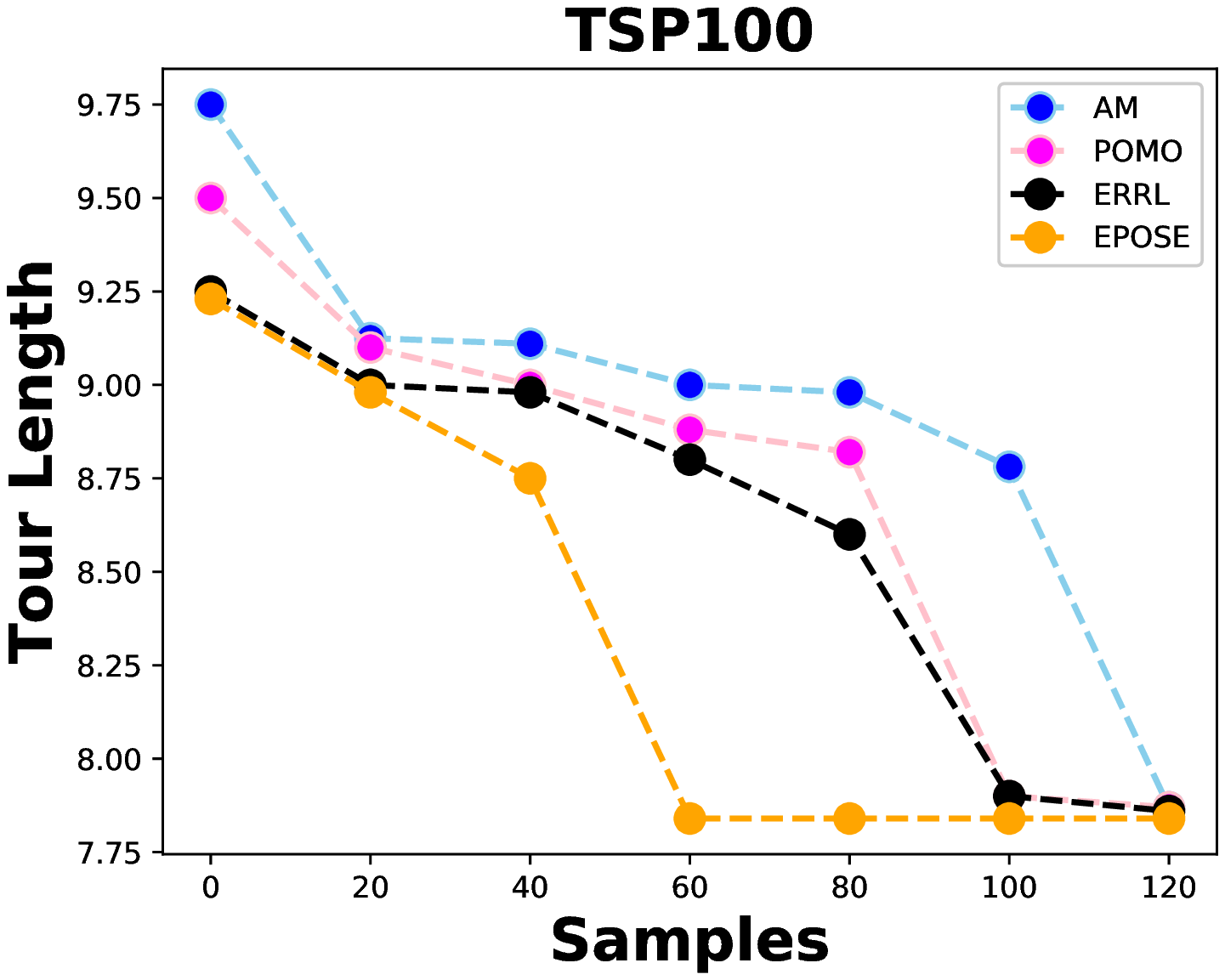}
    \caption{TSP100 sample efficiency}\label{fig:samp}
  \end{subfigure}
  \begin{subfigure}[b]{0.33\columnwidth}
    \includegraphics[width=\linewidth]{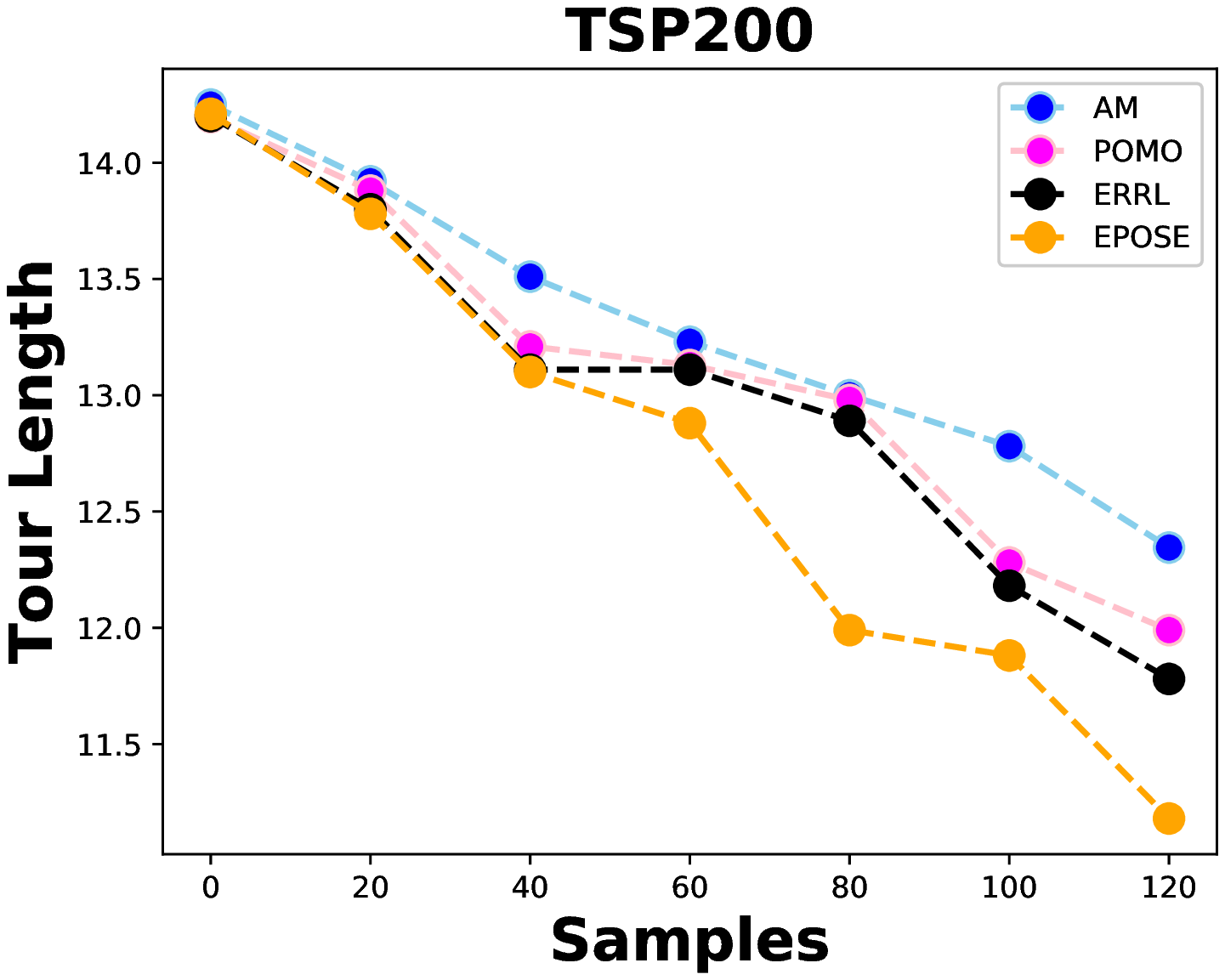}
    \caption{TSP200 sample efficiency}\label{fig:sample}
  \end{subfigure}
   \caption{Sample efficiency analysis for the TSP for various sizes of cities.}\label{fig:search}
\end{center}
\end{figure}

The three existing methods Kool et al.~\cite{kool2018attention}, ERRL~\cite{sultana2021learning} and POMO~\cite{kwon2020pomo} resulted in near-optimal solutions for the routing problems. Therefore, we show the learning curves of TSP50 and TSP100 in Figures~\ref{fig:tsmtime} and~\ref{fig:smalltsp}, which demonstrate that EPOSE learned faster compared to the other three methods. We observed that most of the learning was already completed within 100 epochs for both problems in Figure~\ref{fig:smalltsp}. After each training epoch, we generated 10,000 random instances to use them as a validation set. In Figure~\ref{fig:sam},~\ref{fig:samp} and~\ref{fig:sample}, we illustrate the learning behaviour of all models on the TSP with 50, 100 and 200 cities, respectively. Due to the use of a buffer, the sample efficiency was improved significantly. It is evident from Figure~\ref{fig:search} that our model's learning and convergence were considerably faster.
% Also, most of the learning converge faster than kool et al.~\cite{kool2018attention}, ERRL~\cite{sultana2021learning} and POMO~\cite{kwon2020pomo} models. 

\subsubsection{Ablation study}

In this study, our controlled experiments provide the first principled investigation into the impact of entropy terms with off-policy and on-policy learning techniques, revealing that the learning techniques improve the solution.

Previous ERRL~\cite{sultana2021learning} methods proposed using the on-policy technique along with the entropy term to train models. Our EPOSE uses off-policy learning with the maximum-entropy-based framework technique that has been adopted for other routing problems. We performed additional experiments to train the model with only off-policy learning with a fixed-entropy parameter and off-policy learning with a fixed-entropy parameter. It is evident from Table~\label{tabile:ablation} that the solution quality improved using our EPOSE model, which is based on off-policy learning with maximum entropy that selects entropy parameters automatically.

% In this study, our controlled experiments provide the first principled investigation into the impact of entropy terms with off-policy and on-policy learning techniques, revealing that learning techniques improve the solution.

% Previous ERRL~\cite{sultana2021learning} methods proposed the on-policy technique along with the entropy term to train the model. Our EPOSE used off-policy learning with the maximum entropy-based framework technique that is adopted for other routing problems. We have performed additional experiments to train the model with only off-policy learning fixed entropy parameter and off-policy learning with fixed entropy parameter. The results are given in Table~\label{tabile:ablation} tour length on the methods for TSP and CVRP. It is evident from Table~\label{tabile:ablation} that the solution quality improved using our EPOSE model, which is based on Off-policy learning with maximum entropy that selected entropy parameters automatically. 

\begin{table*}[!ht]
\centering
\caption{Ablation study: tour length of the methods for the TSP and CVRP. ERRL(OPFE): on-policy with fixed entropy. EPOSE (OPFE): off-policy with fixed entropy..}
\scriptsize
%\resizebox{0.4\textwidth}{!}{
\begin{tabular}{|ll|lll|lll|lll|}\hline
%\begin{tabular}{lc|ccc|ccc|ccc}\hline
%\multicolumn{2}{c}{} & \multicolumn{3}{c}{MRPFF=20} & \multicolumn{3}{c}{MRPFF=50} & \multicolumn{3}{c}{MRPFF=100} \\ \hline
%\multicolumn{2}{c|}{Method} & Obj & Gap(\%) & Time(s) & Obj & Gap(\%) & Time(s) & Obj & Gap(\%) & Time(s) \\\hline\hline 
\multicolumn{2}{|c|}{Method} & \multicolumn{3}{|c|}{TSP=20} & \multicolumn{3}{|c|}{TSP=50}& \multicolumn{3}{|c|}{TSP=100} \\ \hline
\textbf{} & & TourL  & Gap(\%) & Time & TourL &  Gap(\%) &  Time& TourL  &   Gap(\%) & Time\\\hline
ERRL(on-policy with fixed entropy) &RL,G& 3.83 &  * &0.1(s) & 5.73 & 0.52 & 1(s)  & 7.85 &1.02 &  3(s)\\
Off-policy with fixed entropy &RL,G& 3.83 &* & 0.1(s) & 5.72&0.35 &  1(s) & 7.80 &0.38  & 3(s)  \\
EPOSE &RL,G& 3.81 &*& 0.1(s) & 5.70 &* &  1(s) & 7.77 & * & 3(s)  \\\hline
\multicolumn{2}{|c|}{Method} & \multicolumn{3}{|c|}{CVRP=20} & \multicolumn{3}{|c|}{CVRP=50}& \multicolumn{3}{|c|}{CVRP=100} \\ \hline
ERRL(on-policy with fixed entropy) &RL,G& 6.34 &  3.25  & 0.1(s) & 10.77 &  3.65& 2(s)  & 16.35 &4.33 &  6(s)\\
Off-policy with fixed entropy &RL,G& 6.30 &2.60 & 0.1(s) & 10.75 &3.46 &  2(s) & 16.20 & 3.38& 6(s)  \\
EPOSE &RL,G& 6.24 &1.62 & 0.1(s) & 10.73 &3.27 &  2(s) & 16.02 & 2.23 & 6(s)  \\\hline
\end{tabular}%}
\label{tabile:ablation}
\end{table*}

\section{{Conclusion and Future Direction}}\label{sec:conclusion}
This work presents EPOSE model, which offers sample efficiency with exploration, and found the model can improve the performance on many routing problems. We demonstrated that using the off-policy-based gradient method can help reinforcement learning-based models prevent premature convergence and accelerate optimisation. The EPOSE algorithm has higher efficiency of sample utilisation compared with the previous state of the art algorithms. We can observe that for all the three routing problems with 20, 50 and 100 nodes, EPOSE consistently outperforms existing state of the art methods. Our algorithms significantly improve the solutions of four routing problems using less training data than existing studies. Moreover, it was found that our framework has linear running time complexity during the testing process.

In this work, we automate the process by reformulating a different maximum entropy reinforcement objective and improved to use of this model in various routing tasks. The model prevents brittleness concerning their hyper-parameters. We found that a compelling choice of exploration strategy requires collecting better samples to produce a better estimation of the gradients. Therefore, exploration helps the model to be sample efficient, which implies better solution quality with less computational time. Moreover, the model is able to balance exploration and exploitation in terms of the task. We want to propose a specialised off-policy gradient-based method combine with search strategies for the routing problems.

\bibliographystyle{unsrt}  
\bibliography{my}


% George Kour and Raid Saabne.
% \newblock Real-time segmentation of on-line handwritten arabic script.
% \newblock In {\em Frontiers in Handwriting Recognition (ICFHR), 2014 14th
%   International Conference on}, pages 417--422. IEEE, 2014.

% George Kour and Raid Saabne.
% \newblock Fast classification of handwritten on-line arabic characters.
% \newblock In {\em Soft Computing and Pattern Recognition (SoCPaR), 2014 6th
%   International Conference of}, pages 312--318. IEEE, 2014.

% \bibitem{hadash2018estimate}
% Guy Hadash, Einat Kermany, Boaz Carmeli, Ofer Lavi, George Kour, and Alon
% \newblock Estimate and replace: A novel approach to integrating deep neural
%   networks with existing applications.
% \newblock {\em arXiv preprint arXiv:1804.09028}, 2018.

% \end{thebibliography}
\newpage
\appendix
\section{Traveling salesman problem}
We need to find the shortest path that visits all n nodes, where the distance between two nodes is the Euclidean distance. The each node location is sampled randomly from the unit square.

\subsection{Graph Attention Model}

The graph attention model used in the EPOSE experiments is the same as that of Kool et al.~\cite{kool2018attention}(which we refer to as "the original AM paper"). Our graph-attention model~\cite{kool2018attention} defines a stochastic policy $p_{\phi}(\pi|I)$ for instance $I$. Based on the probability of chain rule, the selection probability for a sequence $\pi$ can be calculated based on a parameter set $\phi$ of the graph-attention model. Here, the encoder makes embeddings of all input nodes. The decoder produces a permutation $\pi$ of input nodes by generating a node at each time step and masks~\cite{kool2018attention} that node out to prevent the model from revisiting the node. Our encoder is designed based on the graph-attention model~\cite{kool2018attention}. Graph-attention model is a neural network architecture that transmits node information through an attention mechanism.

The decoder uses an attention mechanism similar to Kool et al.~~\cite{kool2018attention} which is based on the decoder part of a transformer model~\cite{vaswani2017attention}. The transformer model is based on a multi-head attention mechanism. However, it cannot directly be applied to solve combinatorial optimisation problems because the output dimension is fixed in advance and cannot be varied according to the input dimension. In contrast, the pointer network~\cite{vinyals2015pointer} uses an attention mechanism to select a member from the input sequence as the output at each decoding step that uses a softmax probability distribution. The pointer network~\cite{vinyals2015pointer} enables a transformer model to apply to combinatorial optimisation problems, where the source sequence determines the length of an output sequence. Our decoder follows the pointer network way to output nodes as a sequence. Each node is related to a probability value as a "pointer" at each decoding time step using the softmax probability distribution. The encoder-decoder attention we discussed later:

\subsubsection{Encoder}
The encoder that we use is similar to the encoder used in Attention Model~\cite{kool2018attention}. From the 2-dimensional input features the encoder computes initial node embeddings $d_h$ which is 128 dimensional. The encoder computes node embedding through a learned linear projection with parameters. The embeddings are updated using the attention layers consisting of two sub-layers. The encoder computes an aggregated embedding of the input graph as the mean of the final node embeddings. The decoder used the input of node embeddings and the graph embedding. The node embedding $h_i$ for $1 \leq i \leq n$ We define $\hbar$ as the mean of all node embeddings.

\subsubsection{Attention Layer}

Following the attention model, each attention layer consists of two sublayers: i.e., Multi-head attention and feed-forward (FF) layer. Multi-head attention executes message passing between the nodes. Feed-forward node-wise fully connected layer. Each sub-layer consists of a skip-connection and batch normalisation. The Multi-head attention sublayer uses 8 heads and the FF sublayer has one hidden (sub)sublayer with dimension 512 and ReLu activation. The attention mechanism by~\cite{vaswani2017attention} is a techniques that passing weighted message between nodes in a graph. The weight of the message value that a node receives from a neighbor depends on the compatibility of its query with the key of the neighbor for details we refer AM~\cite{kool2018attention}. Multi-head attention beneficial to have multiple attention heads. that allows nodes to receive different types of messages from different neighbors. Feed-forward sublayer computes node-wise projections using a hidden sublayer. We use batch normalisation with learnable 128-dimensional affine parameters. The Multi-head attention denoted as $M$ sublayer uses 8 heads with dimensionality $\frac{d_h}{M}=16$, and the FF sublayer has one hidden (sub)sublayer with dimension 512 and ReLu activation.

\subsubsection{Decoder}

The decoder generates sequence at time step $t \in \{1,\cdots,n\}$, based on the embeddings from the encoder and the decoder outputs the node $\pi_{\acute{t}}$ generates at time $\acute{t} < t$. During decoding, augment the graph with a special context node (c) to represent the decoding context. The decoder produces the sequence $\pi$ of input nodes, one node at a time. It takes as input the encoder embeddings (the graph embedding and node embeddings) and a problem specific mask and context. At each time step t, the context consist of the graph embedding and the embeddings of the first and last (previously output) node of the partial tour, where learned placeholders are used if t = 1. Nodes that cannot be visited (since they are already visited) are masked. The decoder context at time t comes from the encoder and the output upto time $t$. for the TSP it consists of the embedding of the graph, the previous (last) node $\pi_{t-1}$ and the first node $\pi_1$.

For TSP, when a partial tour has been constructed, it cannot be changed and the remaining problem is to find a path from the last node, through all the not visited nodes, to the first node. The order and coordinates of other nodes already visited are irrelevant. To know the first and last node, the decoder context consists of embeddings of the first and last node (the context consist of the graph embedding and the embeddings of the first and last (previously output) node of the partial tour). The decoder computes an attention (sub)layer with messages only to the context node. Nodes that cannot be visited are masked. 

\section{Capacitated Vehicle Routing Problem (CVRP)}

The CVRP is a generalisation of the TSP in which case there is a depot and multiple routes should be created, each starting and ending at the depot. In our graph based formulation, we add a special depot node with index 0 and coordinates $m_0$. A vehicle (route) has capacity $C > 0$ and each (regular) node $i \in \{1, \cdots, n\} $ has a demand $0 < \delta_i < C$. Each route starts and ends at the depot and the total demand in each route should not exceed the capacity, so $\sum_{i\in R_j} \delta_i \leq C$, where $R_j$ is the set of node indices assigned to route j. Without loss of generality, we assume a normalised $\hat{C}=1$ as we can use normalised demands $\hat{\delta_i} = \frac{\delta_i}{C}$.

The Split Delivery VRP (SDVRP) is a generalisation of CVRP in which every node can be visited multiple times, and only a subset of the demand has to be delivered at each visit. We follow Kool et al~\cite{kool2018attention} in the instance generation of instances for CVRP20; CVRP50; CVRP100 and normalise the demands by the capacities. The depot location as well as $n$ node locations are sampled uniformly at random in the unit square.

Instances for both CVRP and SDVRP are specified in the same way: an instance with size as a depot location 0, node locations $m_i, i = 1,\cdots,n$ and (normalised) demands $ 0 < \hat{\delta_i} \leq 1$,  $i = 1,\cdots,n$.

Training For the VRP, the length of the output of the model depends on the number of times the depot is visited. In general, the depot is visited multiple times, and in the case of SDVRP also some regular nodes are visited twice. Therefore the length of the solution is larger than n, which requires more memory such that we find it necessary to limit the batch size B to 256 for n = 100. To keep training times tractable and the total number of parameter updates equal, we still process 2500 batches per epoch, for a total of 0.64M training instances per epoch.

\subsection{ATTENTION MODEL FOR THE CVRP}

In order to allow our Attention Model to distinguish the depot node from the regular nodes, we use separate parameters to compute the initial embedding of the depot node. Additionally, we provide the normalised demand as input feature (and adjust the size of parameter accordingly).

Capacity constraints To facilitate the capacity constraints, we keep track of the remaining demands and remaining vehicle capacity, after which they are updated.

The context for the decoder for the VRP at time t is the current/last location and the remaining capacity. Compared to TSP, we do not need placeholders if t = 1 as the route starts at the depot and we do not need to provide information about the first node as the route should end at the depot:

Masking The depot can be visited multiple times, but we do not allow it to be visited at two subsequent time-steps. Therefore, in both layers of the decoder, we change the masking for the depot depends on whether we allow split deliveries. More details we refer to as “the original AM paper”.

\end{document}